\pdfoutput=1

\documentclass{llncs}
\usepackage{amsmath,amssymb} 
\usepackage{cs}
\usepackage{mathsymb}
\usepackage{amsmath}

\usepackage{graphicx}
\usepackage{caption}
\usepackage{color}
\usepackage{subfigure}
\usepackage{times}
\usepackage{algorithm}
\usepackage{algorithmic}
\graphicspath{{./}{./Fig/}{./Figs/}}

\renewcommand{\thesubfigure}{\thefigure.\arabic{subfigure}}
\renewcommand{\thesubfigure}{(\arabic{subfigure})}

\newcommand{\ABTC}{AsymBoost$_{\rm{TC}}$}
\newcommand{\ABTCone}{AsymBoost$_{\rm{TC1}}$}
\newcommand{\ABTCtwo}{AsymBoost$_{\rm{TC2}}$}

\def\logit{\mathrm{logit}}

\def\classifier{F}

\def\sign{ \mathrm{sign} }

\def\argmaxt{\mathrm{argmmax}}
\newcommand{\ADot}{ \ensuremath{  - \,} }

\usepackage[algo2e,vlined,algoruled]{algorithm2e}
\usepackage{algorithm}

\begin{document}

\newcommand{\point}
{
    \raise0.7ex\hbox{.}
}
    
    
\pagestyle{headings}

\mainmatter








\title{Asymmetric Totally-corrective Boosting for Real-time Object Detection}

\author{Peng Wang$^1$\thanks{Work was done when P. W. was visiting NICTA Canberra
Research Laboratory and Australian
National University.}, Chunhua Shen$^{2,3}$, Nick Barnes$^{2,3}$, Hong Zheng$^1$, Zhang Ren$^1$}

%
%

\institute{
Beihang University, Beijing 100191, China
\and
NICTA\thanks{NICTA is funded by the Australian Government's Department of
Communications, Information Technology, and the Arts and the
Australian Research Council through {\em Backing Australia's Ability}
initiative and the ICT Research Center  of Excellence programs.},
Canberra Research Laboratory, Canberra, ACT 2601, Australia 
\and
Australian National University, Canberra, ACT 0200, Australia
}

\maketitle

\begin{abstract}

Real-time object detection is one of the core problems in computer vision.  The cascade boosting
framework proposed by Viola and Jones has become the standard for this problem. In this framework,
the learning goal for each node is asymmetric, which is required to achieve a high detection rate and a
moderate false positive rate.  We develop new boosting algorithms to address this asymmetric learning
problem. 
We show that our methods explicitly optimize asymmetric loss objectives in a totally corrective
fashion.  The methods are totally corrective in the  sense that the coefficients of 
all selected weak classifiers are updated at each iteration. 
In contract, conventional boosting like AdaBoost is stage-wise in that only the current weak classifier's
coefficient is updated.  
At the heart of the totally corrective boosting is the column generation technique.
Experiments on face detection show that our methods outperform the state-of-the-art asymmetric
boosting methods.

\end{abstract}

\section{Introduction}

Due to its important applications in video surveillance, interactive human-machine
interface {\em etc},
real-time object detection has attracted extensive research recently
\cite{paisitkriangkrai2008fast,viola2004robust,viola2002fast,paisitkriangkrai2009cvpr,shen2009onthedual,Shen2010ECCV}.
Although it was introduced a decade ago,
the boosted cascade classifier framework of Viola and Jones \cite{viola2004robust} 
is still considered as the most promising approach for object detection, 
and this framework is the basis which many papers have extended.

One difficulty in object detection is the problem is highly asymmetric.
A common method to detect objects in an image is 
to exhaustively search all sub-windows at all possible scales and positions in the image, 
and use a trained model to detect target objects.
Typically, there are only a few targets in millions of searched sub-windows.
The cascade classifier framework partially solves the asymmetry problem 
by splitting the detection process into several nodes.
Only those sub-windows passing through all nodes are classified as true targets.
At each node, we want to train a classifier with a very high detection rate (\eg, $99.5\%$) 
and a moderate false positive rate (\eg, around $50\%$).
The learning goal of each node should be asymmetric in order to achieve optimal detection
performance. A drawback of standard boosting like AdaBoost in the context of the cascade framework
is it is designed to minimize the overall false rate.
The losses are equal for misclassifying a positive example and a negative example,
which makes it not be able to build an optimal classifier for the 
asymmetric learning goal.

Many subsequent works attempt to improve the performance of object detectors 
by introducing asymmetric loss functions to boosting algorithms.
Viola and Jones proposed asymmetric AdaBoost \cite{viola2002fast}, 
which applies an asymmetric multiplier to one of the classes.
However, this asymmetry is absorbed immediately by the first weak classifier 
because AdaBoost's optimization strategy is  greedy.
In practice, they manually apply the $n$-th root of the multiplier on each iteration 
to keep the asymmetric effect throughout the entire training process. 
Here $ n $ is the number of weak classifiers. 
This heuristic cannot guarantee the solution to be optimal 
and the number of weak classifiers need to be specified before training.
AdaCost presented by Fan {\em et al.} \cite{fan1999adacost} adds a cost adjustment function
on the weight updating strategy of AdaBoost.
They also pointed out that the weight updating rule should consider the 
cost not only on the initial weights but also at each iteration.
Li and Zhang \cite{li2004float} proposed FloatBoost 
to reduce the redundancy of greedy search 
by incorporating floating search with AdaBoost.
In FloatBoost, the poor weak classifiers are deleted when adding the new weak classifier.
Xiao {\em et al.} \cite{rong2003boostingchain} improved the backtrack technique in \cite{li2004float}
and exploited the historical information of preceding nodes into successive node learning. 
Hou {\em et al.} \cite{hou2006learning} used varying asymmetric factors for training different weak
classifiers.
However, because the asymmetric factor changes during training, the loss function remains unclear. 
Pham {\em et al.} \cite{pham08multi} presented a method which trains the asymmetric 
AdaBoost \cite{viola2002fast} classifiers
under a new cascade structure, namely multi-exit cascade.
Like soft cascade \cite{bourdev05softcascade}, 
boosting chain \cite{rong2003boostingchain} and dynamic cascade \cite{rong2007},
multi-exit cascade is a cascade structure which takes the historical information into consideration. 
In multi-exit cascade, the $n$-th node ``inherits'' weak classifiers selected at the preceding $n-1$ nodes.
Wu {\em et al.} \cite{wu2008fast} stated that feature selection and ensemble classifier learning can be
decoupled. They designed a linear asymmetric classifier (LAC) to adjust the linear coefficients
of the selected weak classifiers. 
{K}ullback-{L}eibler Boosting \cite{liu2003kl} iteratively learns robust linear features
by maximizing the Kullback-Leibler divergence.

Much of the previous work is based on AdaBoost
and achieves the asymmetric learning goal by {\em heuristic} weights manipulations
or post-processing techniques.
It is not trivial to assess  how  these heuristics affect the original loss function 
of AdaBoost.
In this work, we construct new boosting algorithms {\em directly} from asymmetric losses.
The optimization process is implemented by column generation.
Experiments on toy data and real data show that our algorithms indeed achieve the asymmetric
learning goal without any heuristic manipulation, and can outperform previous methods.

Therefore, the main contributions of this work are as follows.
\begin{enumerate}
\item
We utilize a general and systematic framework (column generation) to construct new 
asymmetric boosting
algorithms, which can be applied to a variety of asymmetric losses.  There is no heuristic strategy
in our algorithms which may cause suboptimal solutions.  In contrast, 
the global optimal solution is guaranteed for our algorithms.  

Unlike Viola-Jones' asymmetric AdaBoost \cite{viola2002fast},
the asymmetric effect of our methods spreads over the entire
training process.  The coefficients of all weak classifiers are updated at each iteration, which
prevents the first weak classifier from absorbing the asymmetry.  The number of weak
classifiers does not need to be specified before training.

\item
The asymmetric totally-corrective boosting algorithms introduce the asymmetric learning goal into
both feature selection and ensemble classifier learning.  Both the example weights and the linear
classifier coefficients are learned in an asymmetric way. 

\item
In practice, L-BFGS-B \cite{zhu1997lbfgs} is used to solve the primal problem, which runs much
faster than solving the dual problem and also less memory is needed.

\item
We demonstrate that with the totally corrective optimization, 
the linear coefficients of some weak classifiers are set to zero by the algorithm such that
fewer weak classifiers are needed. We present analysis on the theoretical condition and show
how useful the historical information is for the training of successive nodes.

\end{enumerate}

\section{Asymmetric losses}

In this section, we propose two asymmetric losses,
which are motivated by asymmetric AdaBoost \cite{viola2002fast} 
and cost-sensitive LogitBoost \cite{hanmed2010cost},  respectively.

We first introduce an asymmetric cost in the following form:
\begin{align}
\label{EQ:acost}
{\rm ACost} = \left\{ \begin{array}{ll}
                C_1 & \text{ if } y = +1 \text{ and } \sign(\classifier(\bx)) = -1,\\
                C_2 & \text{ if } y = -1 \text{ and } \sign(\classifier(\bx)) = +1, \\
                0   & \text{ if } y = \sign(\classifier(\bx)).
                \end{array}          \right . \notag
\end{align}
Here $ \bx $ is the input data, $ y $ is the label and $ \classifier( \bx) $ is the learned 
classifier. 
Viola and Jones \cite{viola2002fast} 
directly take the product of $\rm ACost$ and the exponential loss $E_{\bX, Y}
[\exp(-y\classifier(\bx)]$ as the asymmetric loss:
\begin{equation}
E_{\bX, Y}  [\big( \bI (y = 1) C_1 + \bI (y = -1) C_2 \big) \exp \big( -y \classifier(\bx) \big)],
\notag
\end{equation}
where $\bI(\cdot)$ is the indicator function.
In a similar manner, we can also form an asymmetric loss from the logistic loss $E_{\bX, Y} [\logit
\big( -y\classifier(\bx) \big)]$:
\begin{equation}
\label{EQ:ALossLogit1}
{\rm ALoss_1} = E_{\bX, Y} [ \big( \bI (y = 1) C_1 + \bI (y = -1) C_2 \big) \logit \big( y
\classifier(\bx) \big) ], 
\end{equation}
where $\logit(x) = \log (1 + \exp(-x))$ is the logistic loss function.  

Masnadi-Shirazi and Vasconcelos
\cite{hanmed2010cost} proposed cost-sensitive boosting algorithms 
which optimize different versions of cost-sensitive losses by the means of gradient descent.
They proved that
the optimal cost-sensitive predictor minimizes the expected loss:
\begin{equation}
-E_{\bX, Y}[\bI (y = 1) \log(\mathrm{p_c} (\bx)) + \bI (y = -1) \log(1 - \mathrm{p_c} (\bx))], \notag
\end{equation}
\begin{equation}
\text{where} \quad \mathrm{p_c} (\bx) = \frac{e^{\gamma \classifier(x) + \eta} }{ e^{\gamma \classifier(x) + \eta} + e^{- \gamma \classifier(x) - \eta}}, \notag
\text{with } \gamma = \frac{C_1 + C_2}{2}, \eta =  \frac{1}{2} \log \frac{C_2}{C_1}. \notag
\end{equation}
With fixing $\gamma$ to $1$, the expected loss can be reformulated to 
\begin{equation}
\label{EQ:ALossLogit2}
{\rm ALoss_2} = E_{\bX, Y} [\logit \big( y \classifier(\bx) + 2 y \eta \big)].
\end{equation}

\section{Asymmetric totally-corrective boosting}

In this section, we construct asymmetric totally-corrective boosting algorithms (termed 
\ABTC \ here) from the losses \eqref{EQ:ALossLogit1} and \eqref{EQ:ALossLogit2} discussed previously.
In contrast to the methods constructing boosting-like algorithms in \cite{hanmed2010cost},
\cite{Friedman2000additivelogistic} and \cite{ratsch2002boostsvm},
we use column generation to design our totally corrective boosting algorithms,
inspired by \cite{demiriz2002lpboost} and \cite{shen2009onthedual}.

Suppose there are $M$ training examples ($M_1$ positives and $M_2$ negatives),
and the sequence of examples are arranged according to the labels (positives first).
The pool $\cH$ contains $N$ available weak classifiers.
The matrix $H \in \mathbb{Z}^{M \times N}$ contains
binary outputs of weak classifiers in $\cH$ for training examples,
namely $H_{ij} = h_j (x_i)$.
We are aiming to learn a linear combination $\classifier_\bw(\cdot) = \sum_{j=1}^{N} w_i h_j (\cdot)$.
$C_1$ and $C_2$ are costs for misclassifying positives and negatives, respectively.
We assign the asymmetric factor $k = C_2 / C_1$ and restrict $\gamma = (C_1 + C_2) / 2$  to $1$,
thus $C_1$ and $C_2$ are fixed for a given $k$.

The problems of the two \ABTC \ algorithms can be expressed as: 
\begin{align}
\label{EQ:primal_asym_logit1}
  \min_{\bw}  \sum_{i=1}^M l_i \logit(z_i) + \theta \b{1}^\T \bw  
  \quad \sst \bw \psd \b{0}, ~ z_i =  y_i H_i \bw, 
\end{align}
where $\bl = [ C_1/M_1, \cdots, C_2/M_2, \cdots ]^{\T}$, and
\begin{align}
\label{EQ:primal_asym_logit2}
  \min_{\bw} \sum_{i=1}^M e_i \logit ( z_i + 2 y_i \eta ) + \theta \b{1}^\T \bw
  \quad \sst \bw \psd \b{0}, ~ z_i =  y_i H_i \bw, 
\end{align}
where $\be = [ 1/M_1, \cdots, 1/M_2, \cdots ]^{\T}$.
In both \eqref{EQ:primal_asym_logit1} and \eqref{EQ:primal_asym_logit2}, 
$z_i$ stands for the margin of the $i$-th training example.
We refer \eqref{EQ:primal_asym_logit1} as \ABTCone \ and \eqref{EQ:primal_asym_logit2} as \ABTCtwo.
Note that here the optimization problems are $ \ell_1 $-norm regularized. 
It is possible to use other format of regularization such as the $ \ell_2$-norm. 

First we introduce a fact that 
the Fenchel conjugate \cite{Boyd2004Convex} of the logistic loss function $\logit(x)$ is 
      \[
            \logit^*( u ) = 
              \begin{cases}
                     (- u ) \log (- u ) + ( 1 + u ) \log ( 1 + u), 
                                 & \!\!\!\!\! 0 \geq u \geq -1;\\
                              \infty,
                                 & \!\!\!\!\! \text{otherwise}.
              \end{cases}
      \]

Now we derive the Lagrange dual \cite{Boyd2004Convex} of \ABTCone.
The Lagrangian of \eqref{EQ:primal_asym_logit1} is 
\begin{equation}
  {L}(\underbrace{\bw,\bz}_\text{primal},\underbrace{\b{\lambda},\bu}_\text{dual}) 
  = \sum_{i=1}^{M} l_i \logit(z_i) + \theta \b{1}^{\T} \bw 
    - \b{\lambda}^{\T} \bw + \sum_{i=1}^{M} u_i (z_i - y_i H_i \bw). \notag
\label{EQ:lagrangian}
\end{equation}
The dual function 
\begin{align}
  {g}(\b{\lambda}, \bu) & = \inf_{\bw, \bz} {L} (\bw, \bz, \b{\lambda}, \bu)  \notag \\
  & = - \sum_{i=1}^{M} \underbrace{ \sup_{z_i} \Big( - u_i z_i - l_i \logit(z_i) \Big)}_
      \text{$l_i \logit^{*} (- u_i / l_i)$}
    + \inf_{\bw} \underbrace{ \Big( \theta \b{1}^{\T} - \b{\lambda}^{\T} - \sum_{i=1}^{M} u_i y_i H_i \Big) }_\text{must be $\b{0}$} \bw. \notag
\end{align}
The dual problem is 
\begin{align}
\label{EQ:dual_asym_logit1}
\max_{\bu} ~ & - \sum_{i=1}^{M} \Big[
u_i \log(u_i) + (l_i - u_i) \log(l_i - u_i) \Big] \notag \\
\sst       ~ & \sum_{i=1}^{M} u_i y_i H_i \nsd \theta \b{1}^{\T},~ 0 \nsd \bu \nsd \bl.  
\end{align}

Since the problem \eqref{EQ:primal_asym_logit1} is convex and the Slater's conditions
are satisfied \cite{Boyd2004Convex},
the duality gap between the primal \eqref{EQ:primal_asym_logit1} and the dual
\eqref{EQ:dual_asym_logit1} is zero.  Therefore, the solutions of \eqref{EQ:primal_asym_logit1} and
\eqref{EQ:dual_asym_logit1} are the same.  
Through the KKT condition, the gradient of Lagrangian \eqref{EQ:lagrangian} 
over primal variable $\bz$ and dual variable $\bu$ should vanish at the optimum.
Therefore, we can obtain the relationship
between the optimal value of $\bz$ and $\bu$:
\begin{align}
u_i^{*} = \frac{ l_i \exp(-z_i^{*})}{1 + \exp(-z_i^{*})}.
\label{EQ:dual_primal1}
\end{align}

Similarly, we can get the dual problem of \ABTCtwo, which is expressed as:
\begin{align}
\label{EQ:dual_asym_logit2}
\max_{\bu} ~ & - \sum_{i=1}^{M} \Big[
u_i \log(u_i) + (e_i - u_i) \log(e_i - u_i) + 2 u_i y_i \eta \Big]
\notag \\
\sst       ~ & \sum_{i=1}^{M} u_i y_i H_i \nsd \theta \b{1}^{\T},~ 0 \nsd \bu \nsd \be, 
\end{align}
with
\begin{align}
 u_i^{*} = \frac{ e_i \exp(-z_i^{*} - 2 y_i \eta)}{1 + \exp(-z_i^{*} - 2 y_i \eta)}.
\label{EQ:dual_primal2}
\end{align}

In practice, the total number of weak classifiers, $N$, could be extremely large, 
so we can not solve the primal problems \eqref{EQ:primal_asym_logit1} and
\eqref{EQ:primal_asym_logit2} directly.  However equivalently, we can optimize the duals
\eqref{EQ:dual_asym_logit1} and \eqref{EQ:dual_asym_logit2} 
iteratively using column generation \cite{demiriz2002lpboost}.
In each round, we add the most violated constraint by finding a weak classifier satisfying:
\begin{equation}
h^{\star} (\cdot) = \argmax_{h(\cdot)} \sum_{i=1}^{M} u_i y_i h(\bx_i).
\label{EQ:weak}
\end{equation} 
This step is the same as training a weak classifier in AdaBoost and LPBoost, 
in which one tries to find a weak classifier with the maximal edge (\ie the minimal weighted error).
The {\em edge} of $h_j$ is defined as $\sum_{i=1}^{M} u_i y_i h_j(x_i)$, 
which is the inverse of the weighted error.
Then we solve the restricted dual problem with one more constraint than the previous round,
and update the linear coefficients of weak classifiers ($\bw$) and the weights of training examples
($\bu$). 
Adding one constraint into the dual problem corresponds to adding one variable into
the primal problem. Since the primal problem and dual problem are equivalent, 
we can either  solve the restricted dual or the restricted primal in practice.
The algorithms of \ABTCone \ and \ABTCtwo \ are summarized in Algorithm~\ref{alg:ABTC}.
Note that, in practice, in order to achieve specific false negative rate (FNR) or false positive rate
(FPR), 
an offset $b$ is needed to be added into the final strong classifier: 
$ \classifier(\bx) = \sum_{j=1}^n w_i h_j (\x) - b $,
which can be obtained by a simple line search.
The new weak classifier $ h'(\cdot) $ corresponds to an extra variable to the primal and an extra
constraint to the dual.  Thus, the minimal value of the primal decreases with growing variables, 
and the maximal value of the dual problem also decreases with growing constraints.
Furthermore, as the optimization problems involved  are  convex, 
 Algorithm~\ref{alg:ABTC} is guaranteed to converge to the
global optimum.

Next we  show  how \ABTC \ introduces the asymmetric learning into feature selection and
ensemble classifier learning.
Decision stumps are the most commonly used type of weak classifiers, and each stump only uses one
dimension of the features.  So the process of training weak classifiers (decision stumps) is equivalent
to feature selection.  In our framework, the weak classifier with the maximum edge (\ie the minimal weighted error) is
selected. 
From \eqref{EQ:dual_primal1} and \eqref{EQ:dual_primal2}, the weight of $i$-th example, namely $u_i$,
is affected by two factors: the asymmetric factor $k$ and the current margin $z_i$.  If we set $k =
1$, the weighting strategy goes back to being symmetric.  On the other hand, the coefficients of
the linear classifier, namely $\bw$, are updated by solving the restricted primal problem at each
iteration.  The asymmetric factor $k$ in the primal is absorbed by all the weak classifiers
currently learned.  So feature selection and ensemble classifier learning both consider the
asymmetric factor $k$.

   \linesnumbered\SetVline
   \begin{algorithm}[t]
   \caption{The training algorithms of \ABTCone \ and \ABTCtwo.}   
   \centering
   \begin{minipage}[]{0.91\linewidth}
   \KwIn{A training set with $M$ labeled examples ($M_1$ positives and $M_2$ negatives);
         termination tolerant $ \varepsilon > 0$;
         regularization parameter $ \theta $;
         asymmetric factor $k$; 
         maximum number of weak classifiers $ N_{\rm max}$.
    }
       { {\bf Initialization}:
            $ N = 0 $;
            $ \bw = {\bf 0} $;
            and $ u_i = l_i/2 \text{ or } e_i/ (1 + k^{-y_i}) $ , $ i = 1$$\cdots$$M$. 
   }

   \For{ $ \mathrm{iteration} = 1 : N_\mathrm{max}$}
   {
     %
     \ADot
        Train a weak classifier $h' (\cdot) = \argmaxt_{h(\cdot)} \sum_{i=1}^{M} u_i y_i h(\bx_i)$.

     \ADot
         Check for the termination condition: \\
         {\bf if}{ $ \mathrm{iteration}  > 1 $ \text{ and } $
                    \sum_{ i=1 }^M  u_i y_i h' ( \x_i )  
                           < \theta + \varepsilon $},
                  { \bf then} break; 

     \ADot  
         Increment the number of weak classifiers
             $N = N + 1$.  

     \ADot
         Add $  h'(\cdot) $ to the restricted master problem;
     %
     %
     %

      \ADot  
         Solve the primal problem \eqref{EQ:primal_asym_logit1} or \eqref{EQ:primal_asym_logit2} 
         (or the dual problem \eqref{EQ:dual_asym_logit1} or \eqref{EQ:dual_asym_logit2}) 
         and update $ u_i$ ($ i = 1\cdots M $) and $w_j$ ($ j = 1\cdots N $) . 
   }
   \KwOut{
         The selected weak classifiers are $ h_1, h_2, \dots, h_N $.
         The final strong classifier is:
         $ F ( \bx ) = \textstyle \sum_{j=1}^{ N } w_j h_j( \bx )$.

   }
   \end{minipage}
   \label{alg:ABTC}
   \end{algorithm}

The number of variables of the primal problem is the number of weak classifiers, 
while for the dual problem, it is the number of training examples. 
In the cascade classifiers for face detection, 
the number of weak classifiers is usually much smaller than the number of training examples,
so solving the primal is much cheaper than solving the dual.
Since the primal problem has  only simple box-bounding constraints,
we can employ L-BFGS-B \cite{zhu1997lbfgs} to solve it.
L-BFGS-B is a tool based on the quasi-Newton method for bound-constrained optimization.

Instead of maintaining the Hessian matrix, 
L-BFGS-B only needs the recent several updates of values and gradients for the cost function
to approximate the Hessian matrix.
Thus, L-BFGS-B requires less memory when running. 
In column generation,
we can use the results from previous iteration as the starting point of current problem,
which leads to further reductions in computation time.

The complementary slackness condition \cite{Boyd2004Convex} suggests that $\lambda_j w_j = 0$.
So we can get the conditions of sparseness: 
\begin{align}
\text{If } \lambda = \theta - \textstyle \sum_{i=1}^{M} u_i y_i H_{i,j} > 0, \text{then } w_j = 0.
\label{EQ:sparse}
\end{align}
This means that, if the weak classifier $h_j(\cdot)$ 
is so ``weak'' that its edge is less than $\theta$ under the current distribution $\bu$, 
its contribution to the ensemble classifier is  ``zero''.
From another viewpoint, the $\ell_1$-norm regularization term in the primal 
\eqref{EQ:primal_asym_logit1} and \eqref{EQ:primal_asym_logit2},
leads to a sparse result.
The parameter $\theta$ controls the degree of the sparseness.
The larger $\theta$ is, the sparser the result would  be.

\section{Experiments}

\subsection{Results on synthetic data}
To show the behavior of our algorithms, we construct a $2$D data set, 
in which the positive data follow the $2$D normal distribution ($N(0, 0.1\I)$),
and the negative data form a ring with uniformly distributed angles and normally 
distributed radius ($N(1.0, 0.2)$).
Totally $2000$ examples are generated ($1000$ positives and $1000$ negatives), 
$50\%$ of data for training and the other half for test.
We compare AdaBoost, \ABTCone \ and \ABTCtwo \ on this data set.
All the training processes are stopped at $100$ decision stumps.
For \ABTCone \ and \ABTCtwo, we fix $\theta$ to $0.01$, 
and use a group of $k$'s $\{1.2, 1.4, 1.6, 1.8, 2.0, 2.2, 2.4, 2.6, 2.8, 3.0\}$.

\begin{figure}[h!]
         \centering
         \subfigure[\ABTCone \ vs AdaBoost]{
            \includegraphics[width=0.48\textwidth]{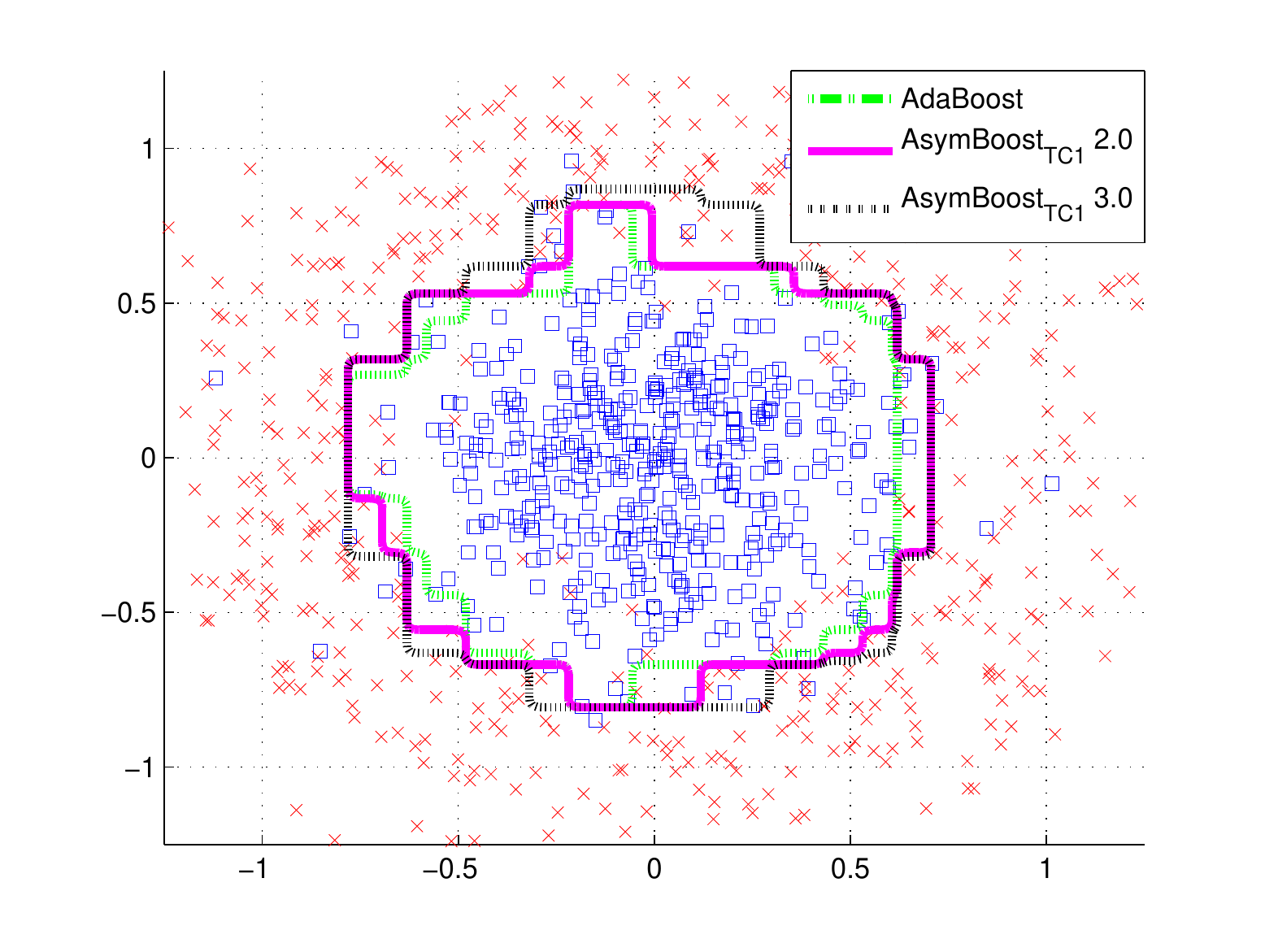}
         }
         %
         \subfigure[\ABTCtwo \ vs AdaBoost]{
            \includegraphics[width=0.48\textwidth]{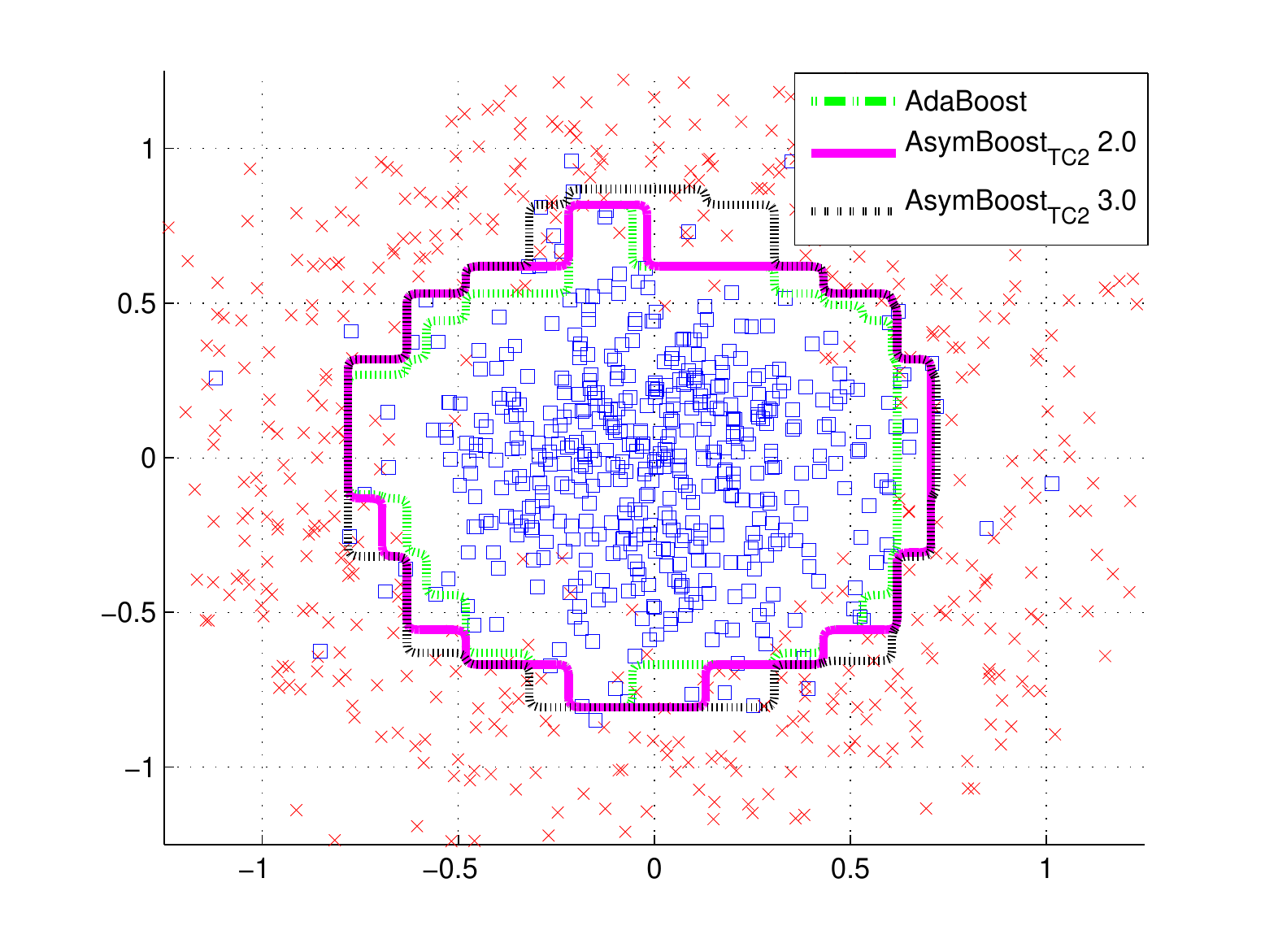}
         }
         %
         \subfigure[False rates for \ABTCone]{
            \includegraphics[width=0.48\textwidth]{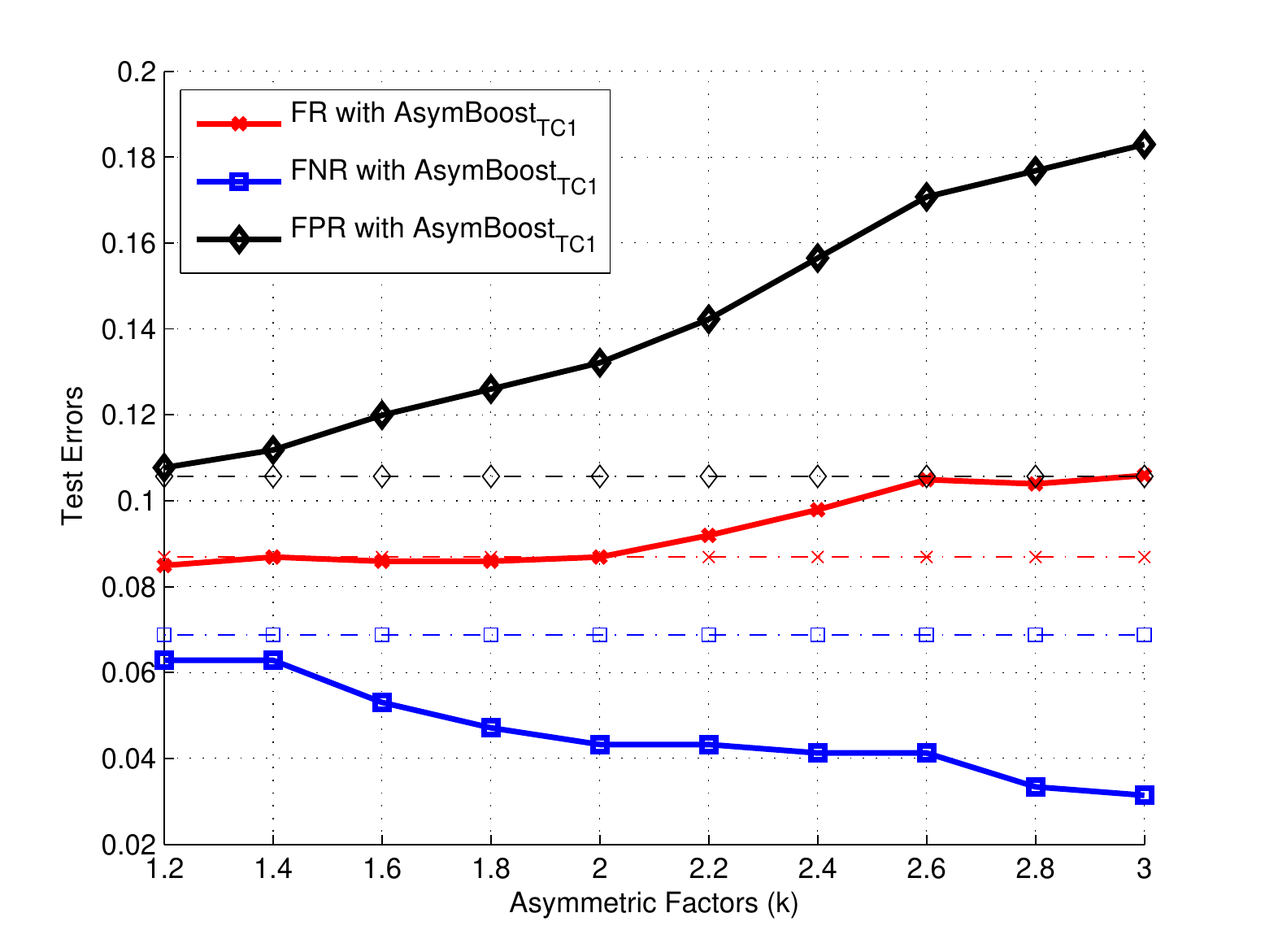}
         }
         %
         \subfigure[False rates for \ABTCtwo]{
            \includegraphics[width=0.48\textwidth]{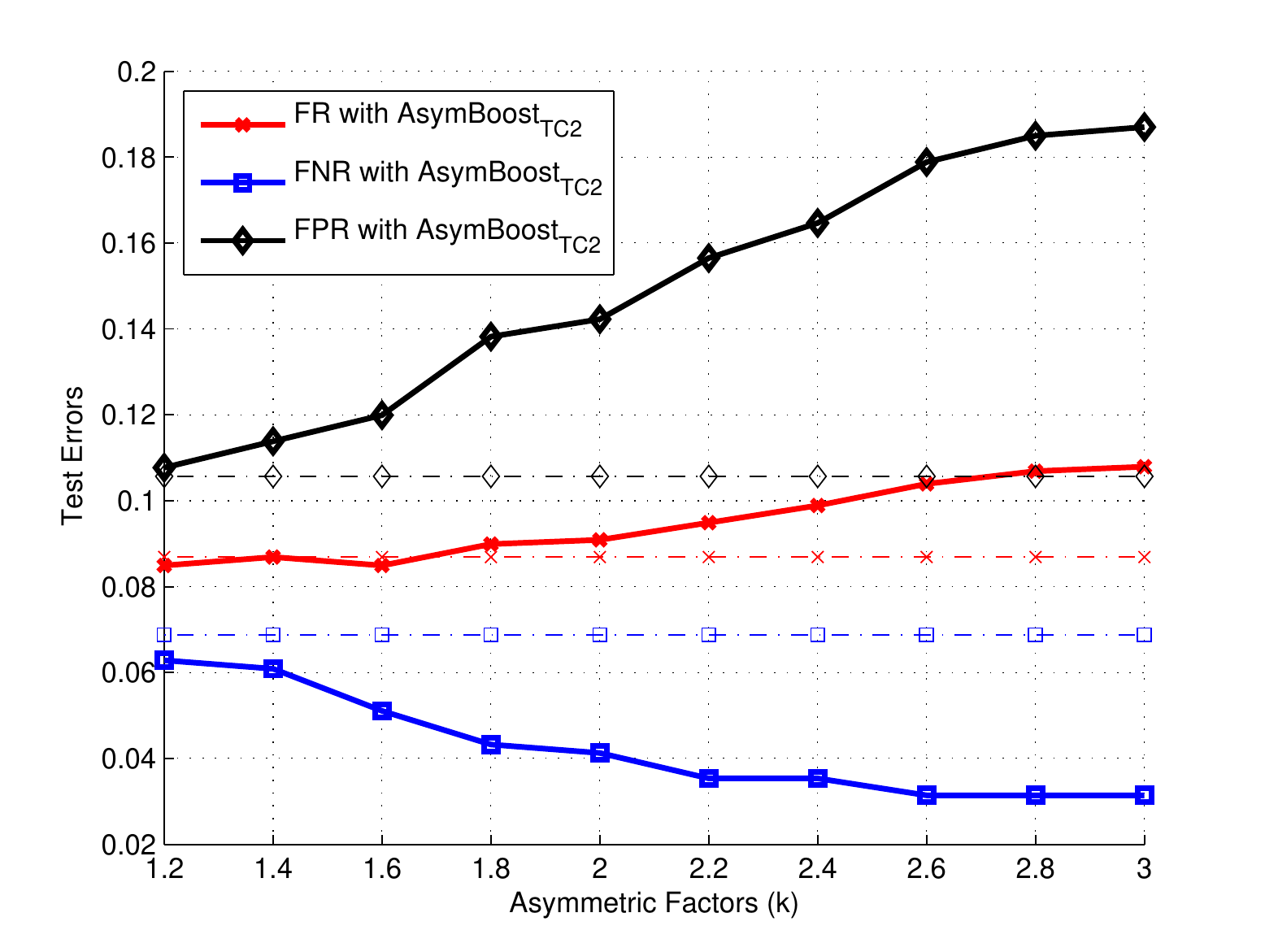}
         }
         \caption{Results on the synthetic data for \ABTCone \ and \ABTCtwo, with a group of asymmetric factor $k$s.
                  As the baseline, the results for AdaBoost are also shown in these figures.
                  ($1$) and ($2$) demonstrate decision boundaries learned by \ABTCone \ and \ABTCtwo, with $k$ is $2.0$ or $3.0$.
                  The {\color{red}$\times$}'s and {\color{blue}$\square$}'s stand for training negatives and training positives respectively.
                  ($3$) and ($4$) demonstrate false rates (FR), false positive rates (FPR) and false negative rates (FNR) on test set 
                  with a group of $k$s ($1.2, 1.4, 1.6, 1.8, 2.0, 2.2, 2.4, 2.6, 2.8$ or $3.0$),
                  and the corresponding rates for AdaBoost is shown as dashed lines.}
                
         \label{FIG:toy}
         \end{figure}

From Figures~\ref{FIG:toy} $(1)$ and $(2)$, 
we find that the larger $k$ is, the bigger the area for positive output becomes, 
which means that the asymmetric LogitBoost tends to make a positive decision 
for the region where positive and negative data are mixed together.
Another observation is that \ABTCone \ and \ABTCtwo \
have almost the same decision boundaries on this data set with same $k$'s.

Figures~\ref{FIG:toy} $(3)$ and $(4)$ demonstrate trends of false rates with the growth of asymmetric
factor ($k$).  The results of AdaBoost is considered as the baseline.  For all $k$'s, 
\ABTCone \ and \ABTCtwo \ achieve lower false negative rates and higher false positive
rates than AdaBoost.  With the growth of $k$, \ABTCone \ and \ABTCtwo \ become more aggressive to reduce the false
negative rate, with the sacrifice of a higher false positive rate.

\subsection{Face detection}
We collect $9832$ mirrored frontal face images and about $10115$ large background images.
$5000$ face images and $7000$ background images are used for training, 
and $4832$ face images and $3115$ background images for validation.
Five basic types of Haar features are calculated on each $24 \times 24$ image, and 
totally generate $162336$ features.
Decision stumps on those $162336$ features construct the pool of weak classifiers.

\textbf{Single-node detectors} 
Single-node classifiers with AdaBoost, \ABTCone \ and \ABTCtwo \ are trained.
The parameters $\theta$ and $k$ are simply set to $0.001$ and $7.0$.
$5000$ faces and $5000$ non-faces are used for training, while $4832$ faces and $5000$ non-faces are used for test.
The training/validation non-faces are randomly cropped from training/validation background images. 

\begin{figure}[t]
         \begin{center}
         \subfigure[DR with fixed FPR]{
            \includegraphics[width=0.75\textwidth]{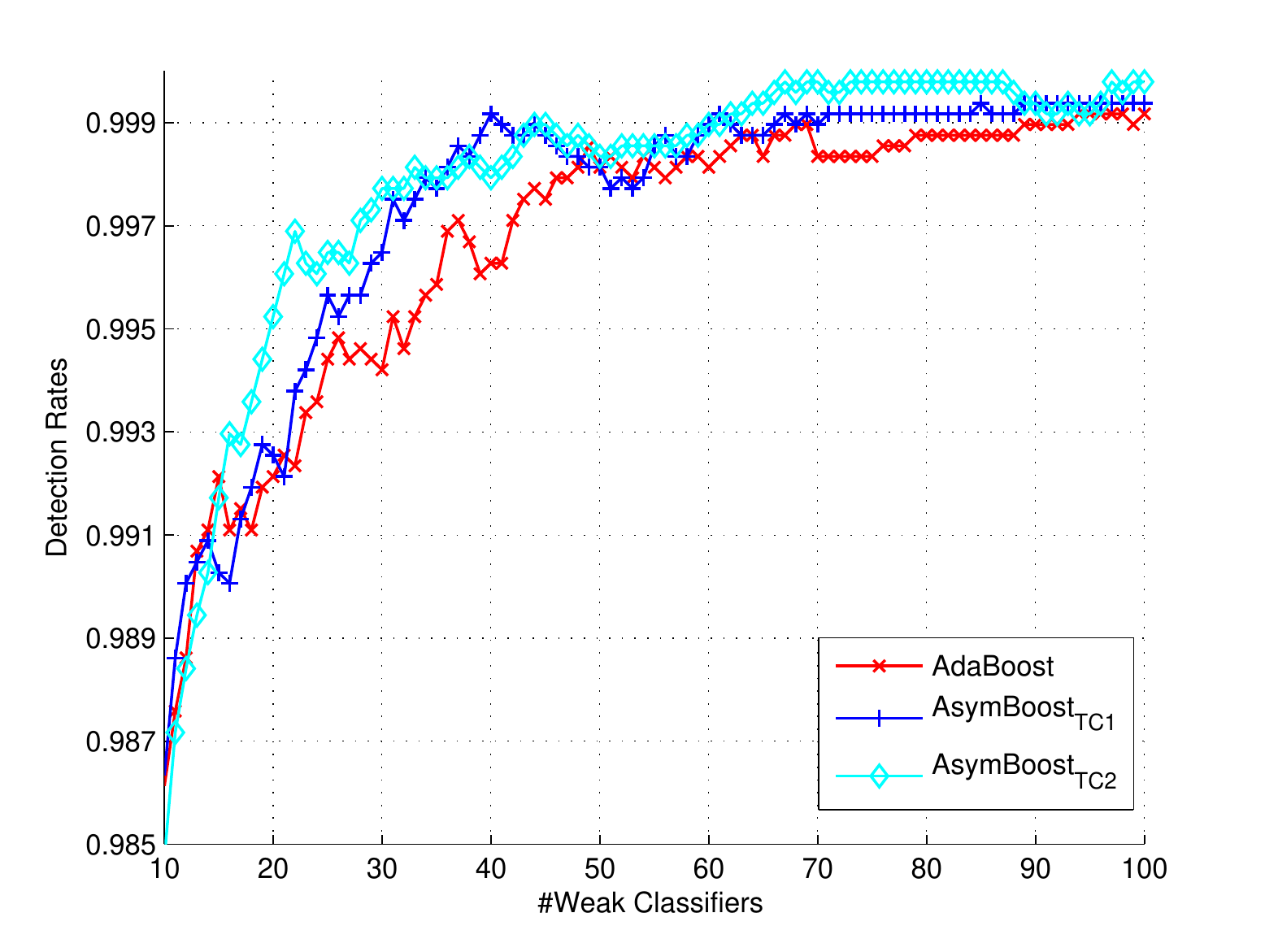}
         }
         %
         \subfigure[FPR with fixed DR]{
            \includegraphics[width=.75\textwidth]{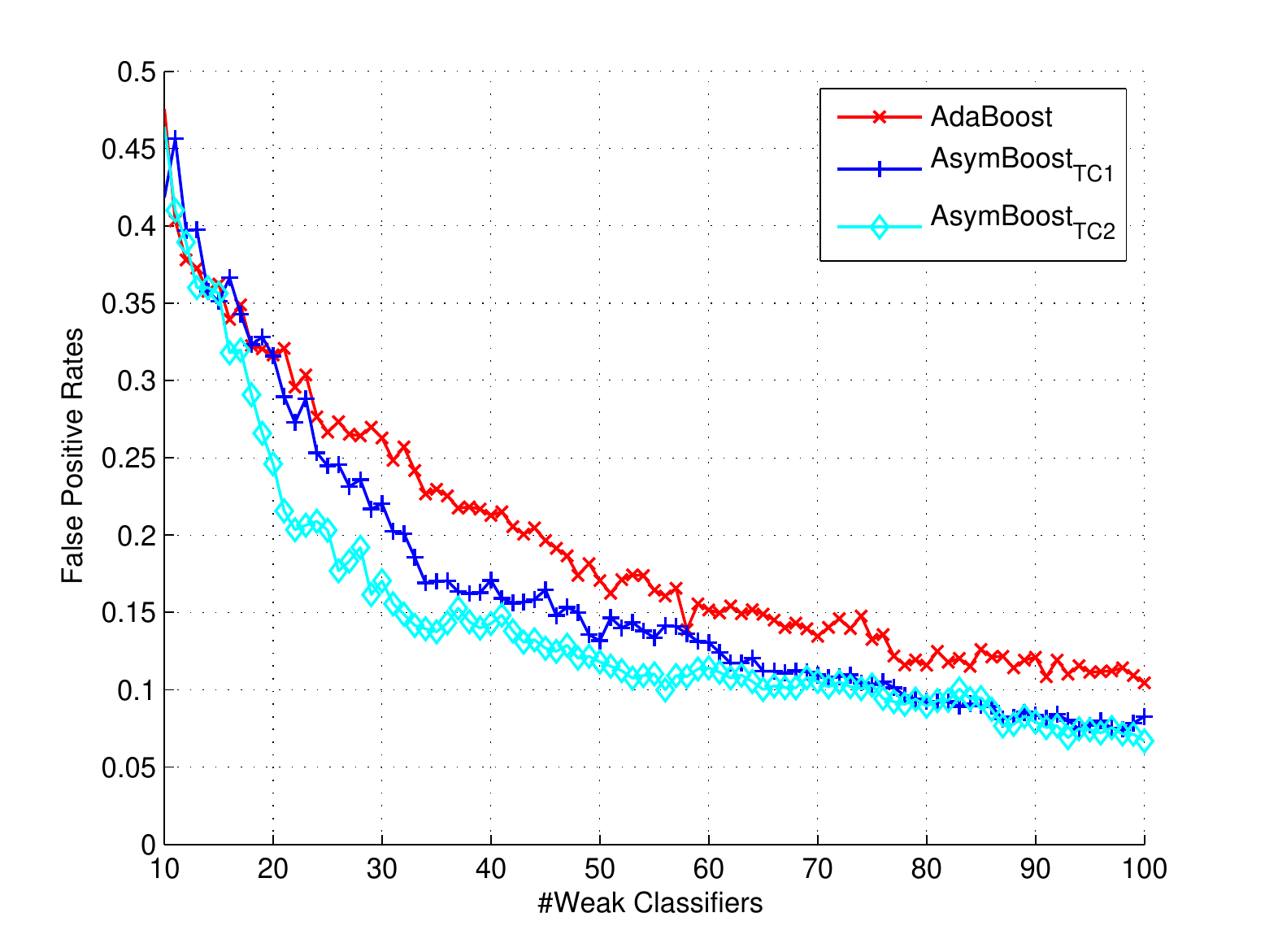}
         }
         \end{center}

         \caption{Testing curves of single-node classifiers for AdaBoost, \ABTCone \ and \ABTCtwo.
                  All the classifiers use the same training and test data sets.
                  ($1$) shows curves of detection rates (DR) with false positive rates (FPR) fixed to $0.25$,
                  ($2$) shows curves of FPR with DR fixed to $0.995$.
                  FPR or DR are evaluated at each weak classifier.
                 }
         \label{FIG:1node}
         \end{figure}

Figure~\ref{FIG:1node} $(1)$ shows curves of detection rate with the false positive rate fixed at $0.25$,
while curves of false positive rates with $0.995$ detection rate are shown in Figure~\ref{FIG:1node} $(2)$.
We set the false positive rate fixed to $0.25$ rather than the commonly used $0.5$ 
in order to slow down the increasing speed of detection rates,
otherwise detection rates would converge to $1.0$ immediately.  
The increasing/decreasing speed of detection rate/false positive rate is faster than 
reported in \cite{li2004float} and \cite{rong2003boostingchain}.
The reason is possibly that we use $10000$ examples for training and $9832$ for testing,
which are smaller than the data used in \cite{li2004float} and \cite{rong2003boostingchain}
($18000$ training examples and $15000$ test examples).
We can see that under both situations, our algorithms achieve better performances 
than AdaBoost in most cases.

The benefits of our algorithms can be expressed in two-fold:
  (1) Given the same learning goal, our algorithms tend to use smaller number of weak classifiers.
   For example, from Figure~\ref{FIG:1node} $(2)$, 
   if we want a classifier with a $0.995$ detection rate and a $0.2$ false positive rate,
   AdaBoost needs at least $43$ weak classifiers while \ABTCone \ needs $32$ and \ABTCtwo \ needs only $22$.
  (2) Using the same number of weak classifiers, our algorithms achieve a higher detection rate or a lower false positive rate.
   For example, from Figure~\ref{FIG:1node} $(2)$, using $30$ weak classifiers, 
   both \ABTCone \ and \ABTCtwo \ achieve higher detection rates ($0.9965$ and $0.9975$) than AdaBoost ($0.9945$).

\textbf{Complete detectors}
Secondly, we train complete face detectors with AdaBoost, asymmetric-AdaBoost, \ABTCone \ and \ABTCtwo.
All detectors are trained using the same training set. 
We use two types of cascade framework for the detector training: 
the traditional cascade of Viola and Jones \cite{viola2004robust} and the multi-exit cascade presented in \cite{pham08multi}.
The latter utilizes decision information of previous nodes when judging instances in the current node.
For fair comparison, all detectors use $24$ nodes and $3332$ weak classifiers.
For each node, $5000$ faces + $5000$ non-faces are used for training, 
and $4832$ faces + $5000$ non-faces are used for validation.   
All non-faces are cropped from background images. 
The asymmetric factor $k$ for asymmetric-AdaBoost, \ABTCone \ and \ABTCtwo \ are 
selected from $\{1.2,$ $ 1.5, 2.0, 3.0, 4.0, 5.0, 6.0\}$.
The regularization factor $\theta$ for \ABTCone \ and \ABTCtwo \ are chosen from 
$\{\frac{1}{50}, $ $\frac{1}{60}, $ $\frac{1}{70}, $ $
\frac{1}{80},$ $ \frac{1}{90}, $ $\frac{1}{100},$ $ \frac{1}{200},$ $ \frac{1}{400}, $ $
\frac{1}{800},$ $ \frac{1}{1000}\}$.
It takes about four hours to train a \ABTC \ face detector 
on a machine with $8$ Intel Xeon E5520 cores and $32$GB memory.
Comparing with AdaBoost, 
only around $0.5$ hour extra time is spent on solving the primal problem at each iteration.
We can say that, in the context of face detection, 
the training time of \ABTC \ is nearly the same as AdaBoost. 
 
\begin{figure}[t]
         \begin{center}
         \subfigure[]{
            \includegraphics[width=0.75\textwidth]{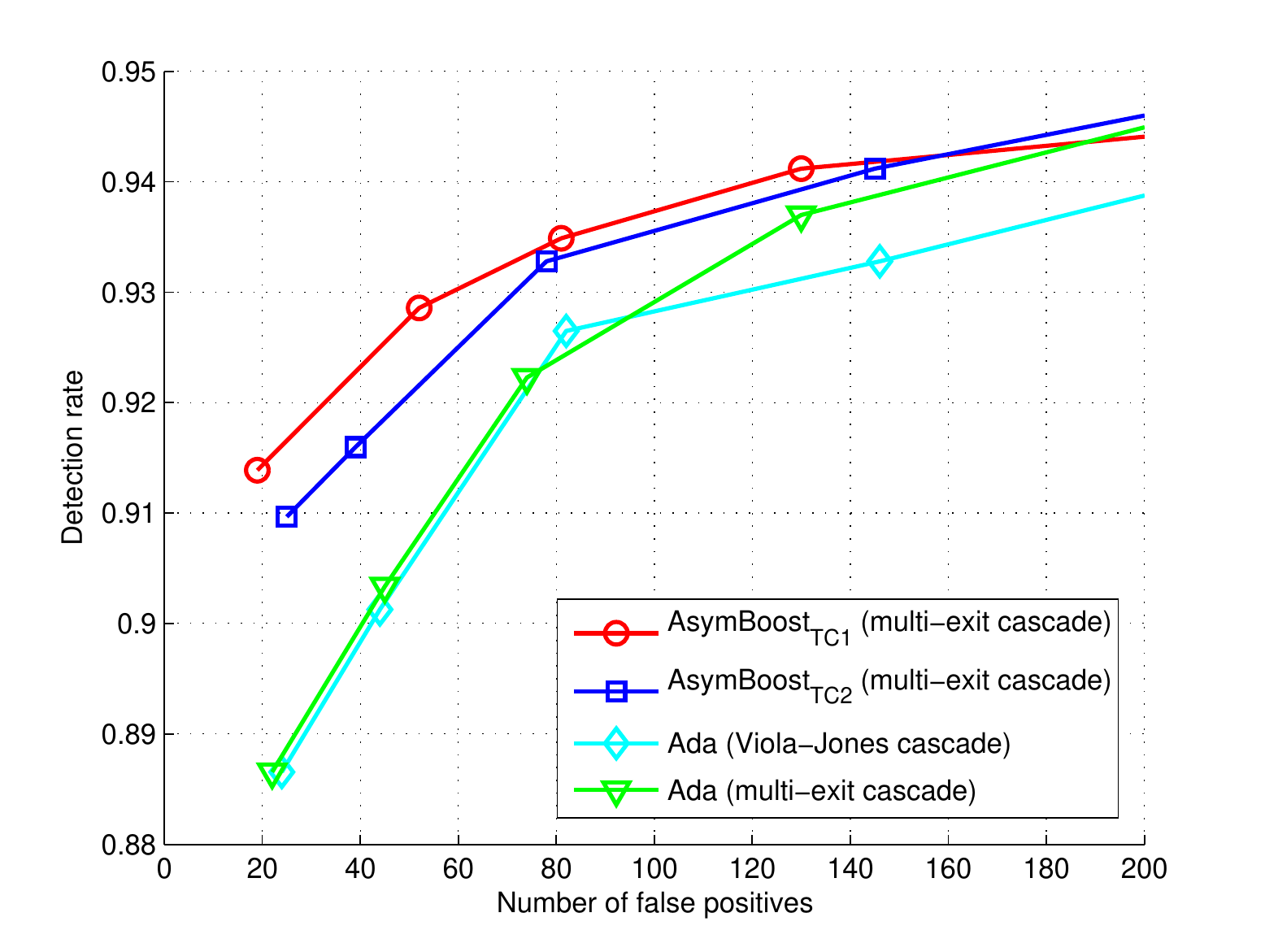}
         }
         %
         \subfigure[]{
            \includegraphics[width=0.75\textwidth]{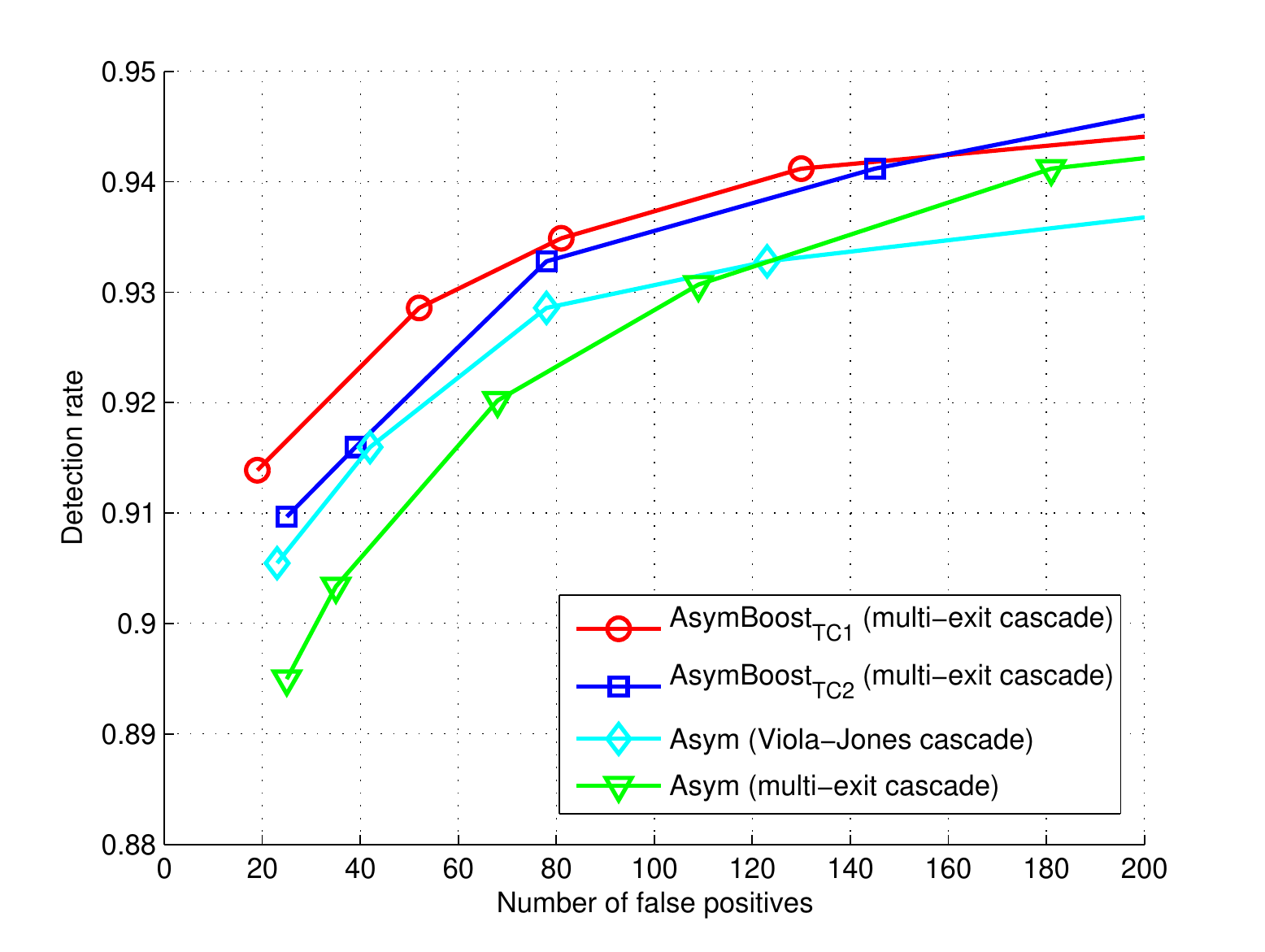}
         }
         \end{center}
         \caption{Performances of cascades evaluated by ROC curves on the MIT+CMU data set.
                  AdaBoost is referred to ``Ada'', and Asymmetric AdaBoost \cite{viola2004robust} is referred to ``Asym''.
                  ``Viola-Jones cascade'' means the traditional cascade used in \cite{viola2002fast} .
                 }
         \label{FIG:roc}
         \end{figure}

ROC curves on the CMU/MIT data set are shown in Figure~\ref{FIG:roc}. 
Those images containing ambiguous faces are removed and $120$ images are retained.  
From the figure, we can see that, asymmetric-AdaBoost outperforms AdaBoost in both Viola-Jones cascade and multi-exit cascade, 
which coincide with what reported in \cite{viola2002fast}. 
Our algorithms have better performances than all other methods in all points
and the improvements are more significant when the false positives are less than $100$,
which is the most commonly used region in practice.

As mentioned in the previous section, our algorithms produce sparse results to some extent.
Some linear coefficients are zero when the corresponding weak classifiers satisfy the condition \eqref{EQ:sparse}.
In the multi-exit cascade, the sparse phenomenon becomes more clear. 
Since correctly classified negative data are discarded after each node is trained, 
the training data for each node are different.
The ``closer'' nodes share more common training examples, while the nodes ``far away'' from each other have distinct training data.
The greater the distance between two nodes, the more uncorrelated they become.
Therefore, the weak classifiers in the early nodes may perform poorly on the last node, thus tending to obtain zero coefficients.
We call those weak classifiers with non-zero coefficients ``effective'' weak classifiers.
Table~\ref{tab:weak_num} shows the ratios of ``effective'' weak classifiers contributed by one node to a specific successive node.
To save space, only the first $15$ nodes are demonstrated.
We can see that, the ratio decreases with the growth of the node index,           
which means that the farther the preceding node is from the current node, the less useful it is for the current node.
For example, the first node has almost no contribution after the eighth node. 
Table~\ref{tab:weak_num2} shows the number of effective weak classifiers used 
by our algorithm and the traditional stage-wise boosting.
All weak classifiers in stage-wise boosting have non-zero coefficients, 
while our totally-corrective algorithm uses much less effective weak classifiers.

         \begin{table*}[t]
         \centering
         \caption
         {
The ratio of weak classifiers selected at the $i$-th node (column) appearing with non-zero coefficients in the $j$-th node (row).  
          The ratios decrease along with the growth of the node index in each column. 
         }
         \begin{tabular}{|l|c|c|c|c|c|c|c|c|c|c|c|c|c|c|c|}
         \hline\hline
         $ \text{Node Index} $                               & 1  & 2  & 3 & 4 & 5 & 6 & 7 & 8 & 9 & 10 & 11 & 12 & 13 & 14 & 15     \\
     \hline
{ 1} & $1.00$  &         &         &         &         &         &         &         &         &         &         &         &         &         &         \\ 
{ 2} & $1.00$  & $1.00$  &         &         &         &         &         &         &         &         &         &         &         &         &         \\ 
{ 3} & $1.00$  & $1.00$  & $1.00$  &         &         &         &         &         &         &         &         &         &         &         &         \\ 
{ 4} & $0.86$  & $1.00$  & $0.97$  & $1.00$  &         &         &         &         &         &         &         &         &         &         &         \\ 
{ 5} & $0.43$  & $0.93$  & $0.97$  & $0.97$  & $1.00$  &         &         &         &         &         &         &         &         &         &         \\ 
{ 6} & $0.71$  & $0.93$  & $0.90$  & $1.00$  & $0.96$  & $1.00$  &         &         &         &         &         &         &         &         &         \\ 
{ 7} & $0.43$  & $0.87$  & $0.87$  & $0.97$  & $0.92$  & $0.92$  & $1.00$  &         &         &         &         &         &         &         &         \\ 
{ 8} & $0.29$  & $0.40$  & $0.70$  & $0.73$  & $0.74$  & $0.88$  & $0.74$  & $1.00$  &         &         &         &         &         &         &         \\ 
{ 9} & $0.00$  & $0.27$  & $0.50$  & $0.60$  & $0.76$  & $0.72$  & $0.66$  & $0.67$  & $1.00$  &         &         &         &         &         &         \\ 
{10} & $0.14$  & $0.27$  & $0.43$  & $0.60$  & $0.62$  & $0.70$  & $0.62$  & $0.66$  & $0.60$  & $1.00$  &         &         &         &         &         \\ 
{11} & $0.00$  & $0.20$  & $0.33$  & $0.50$  & $0.52$  & $0.54$  & $0.60$  & $0.59$  & $0.56$  & $0.48$  & $1.00$  &         &         &         &         \\ 
{12} & $0.14$  & $0.20$  & $0.40$  & $0.40$  & $0.56$  & $0.50$  & $0.54$  & $0.61$  & $0.55$  & $0.46$  & $0.36$  & $1.00$  &         &         &         \\ 
{13} & $0.00$  & $0.13$  & $0.33$  & $0.37$  & $0.36$  & $0.54$  & $0.40$  & $0.47$  & $0.47$  & $0.46$  & $0.43$  & $0.25$  & $1.00$  &         &         \\ 
{14} & $0.00$  & $0.07$  & $0.17$  & $0.40$  & $0.28$  & $0.50$  & $0.42$  & $0.49$  & $0.50$  & $0.53$  & $0.45$  & $0.43$  & $0.35$  & $1.00$  &         \\ 
{15} & $0.00$  & $0.13$  & $0.20$  & $0.27$  & $0.36$  & $0.38$  & $0.46$  & $0.41$  & $0.52$  & $0.42$  & $0.49$  & $0.44$  & $0.34$  & $0.27$  & $1.00$  \\ 
         \hline\hline
         \end{tabular}
         \label{tab:weak_num}
         \end{table*}

 
 
         \begin{table*}[t]
         \centering
         \caption
         { Comparison of the numbers of the effective weak classifiers for the stage-wise boosting (SWB) and the totally-corrective boosting (TCB). 
           We take AdaBoost and \ABTCone \ as representative types of SWB and TCB, both of which are trained in the multi-exit cascade for face detection.
         }
         \begin{tabular}{|l|c|c|c|c|c|c|c|c|c|c|c|c|c|c|c|c|c|c|}
         \hline\hline
         $ \text{Node Index} $  & 1  & 2  & 3 & 4 & 5 & 6 & 7 & 8 & 9 & 10 & 11 & 12 & 13 & 14 & 15 & 16 &17 & 18    \\
     \hline
{SWB} & $7$  & $22$  & $52$  & $82$  & $132$  & $182$  & $232$  & $332$  & $452$  & $592$  & $752$  & $932$  & $1132$  & $1332$  & $1532$ & $1732$ & $1932$ & $2132$       \\ 
{TCB} & $7$  & $22$  & $52$  & $80$  & $125$  & $174$  & $213$  & $269$  & $331$  & $441$  & $464$  & $538$  & $570$   & $681$   & $717$  & $744$  & $742$  & $879$      \\ 
         \hline\hline
         \end{tabular}
         \label{tab:weak_num2}
         \end{table*}

\section{Conclusion}
We have proposed two asymmetric totally-corrective boosting algorithms for object detection, which
are implemented by the column generation technique in convex optimization.  Our algorithms introduce
asymmetry into both feature selection and ensemble classifier learning in a systematic
way.

Both our algorithms achieve better results for face detection than AdaBoost and 
Viola-Jones' asymmetric AdaBoost.
An observation is that we can not see great differences 
on performances between \ABTCone \ and \ABTCtwo \ in our experiments.
For the face detection task, AdaBoost already achieves a very promising result, 
so the improvements of our method are not very significant.

One drawback of our algorithms is there are two parameters to be tuned. 
For different nodes, the optimal parameters should not be the same.
In this work, we have used the same parameters for all nodes.
Nevertheless, since the probability of negative examples decreases with the node index, 
the degree of the asymmetry between positive and negative examples also deceases.
The optimal $k$ may decline with the node index.

The framework for constructing totally-corrective boosting algorithms is general, 
so we can consider other asymmetric losses (\eg, asymmetric exponential loss) 
to form new asymmetric boosting algorithms. 
In column generation, there is no restriction that only one constraint is added
at each iteration.
Actually, we can add several violated constraints at each iteration, 
which means that we can produce multiple weak classifiers in one round.
By doing this, we can speed up the learning process.

Motivated by the analysis of sparseness, we find that the very early nodes contribute little
information for training the later nodes.  Based on this, we can exclude some useless nodes when
the node index grows, which will simplify the multi-exit structure and shorten the testing time.

\bibliographystyle{splncs}
\bibliography{asymboosttc}

\end{document}